\documentclass[10pt,twocolumn,letterpaper]{article}

\usepackage[pagenumbers]{cvpr} 

\usepackage{graphicx}
\usepackage{amsmath}
\usepackage{amssymb}
\usepackage{booktabs}
\usepackage{tabulary,multirow,overpic,xcolor}   
\usepackage{makecell}
\usepackage{soul}
\usepackage{float}
\usepackage{enumitem}

\makeatletter
\AfterEndEnvironment{algorithm}{\let\@algcomment\relax}
\AtEndEnvironment{algorithm}{\kern2pt\hrule\relax\vskip3pt\@algcomment}
\let\@algcomment\relax
\newcommand\algcomment[1]{\def\@algcomment{\footnotesize#1}}
\renewcommand\fs@ruled{\def\@fs@cfont{\bfseries}\let\@fs@capt\floatc@ruled
  \def\@fs@pre{\hrule height.8pt depth0pt \kern2pt}%
  \def\@fs@post{}%
  \def\@fs@mid{\kern2pt\hrule\kern2pt}%
  \let\@fs@iftopcapt\iftrue}
\makeatother

\usepackage{algorithm}
\usepackage{listings}

\usepackage[pagebackref,breaklinks,colorlinks,citecolor=brown]{hyperref}

\usepackage[capitalize]{cleveref}
\crefname{section}{Sec.}{Secs.}
\Crefname{section}{Section}{Sections}
\Crefname{table}{Table}{Tables}
\crefname{table}{Tab.}{Tabs.}

\newlength\savewidth\newcommand\shline{\noalign{\global\savewidth\arrayrulewidth
  \global\arrayrulewidth 1pt}\hline\noalign{\global\arrayrulewidth\savewidth}}

\newcommand{\statement}[1]{\vspace{1mm}\noindent\textbf{#1}}

\def\cvprPaperID{*****} 
\def\confName{CVPR}
\def\confYear{2022}

\begin{document}

\title{Large-Scale Product Retrieval with Weakly Supervised Representation Learning}

\author{Xiao Han\thanks{Equal contributions.}\quad Kam Woh Ng\footnotemark[1]\quad Sauradip Nag\quad Zhiyu Qu \\
Centre for Vision, Speech and Signal Processing (CVSSP), University of Surrey \\
iFlyTek-Surrey Joint Research Centre on Artificial Intelligence \\
{\tt\small \{xiao.han, kamwoh.ng, s.nag, z.qu\}@surrey.ac.uk} \\
{\tt\small \href{https://github.com/01BB01/eBayChallenge}{[Code]} \quad \href{https://wandb.ai/01bb01/fgvc9_ebay_challenge}{[Logs]} \quad \href{https://docs.google.com/spreadsheets/d/1hVdafVw6WkXVwJSrfg6SzkQQd2HZPnPgDGR5EX_Ilb0/edit?usp=sharing}{[Docs]}} 
}
\maketitle

\begin{abstract}
Large-scale weakly supervised product retrieval is a practically useful yet computationally challenging problem.
This paper introduces a novel solution for the eBay Visual Search Challenge (eProduct) held at the Ninth Workshop on Fine-Grained Visual Categorisation workshop (FGVC9) of CVPR 2022. 
This competition presents two challenges: {\bf(a)} E-commerce is a drastically fine-grained domain including many products with subtle visual differences; {\bf(b)} A lacking of target instance-level labels for model training, with only coarse category labels and product titles available.
To overcome these obstacles, we formulate a strong solution by a set of dedicated designs:
(a) Instead of using text training data directly, we mine thousands of pseudo-attributes from product titles and use them as the ground truths for multi-label classification. 
(b) We incorporate several strong backbones with advanced training recipes for more discriminative representation learning.
(c) We further introduce a number of post-processing techniques including whitening, re-ranking and model ensemble
for retrieval enhancement.
By achieving 71.53\% MAR, our solution \texttt{Involution King} achieves the {\bf \em second} position on the leaderboard.

\end{abstract}

\section{Introduction}
\label{sec:intro}

The transition from offline to online shopping 
has been wide and deep across every aspect of our life.
In 2020, retail e-commerce sales worldwide amounted to 4.28 trillion US dollars and are projected to grow to 5.4 trillion US dollars in 2022.
Under this context, 
large-scale product identification becomes a major challenge 
for online service platforms.
A powerful product search system can improve product discoverability and reachability, seller-buyer engagement and conversion rate in e-commerce \cite{yuan2021eproduct,yu2022commercemm,han2022fashionvil}.

To solve this grant challenge, eBay, a world-leading e-commerce company, introduced a million-scale benchmark, namely eProduct \cite{yuan2021eproduct} to foster more advanced AI techniques.
We participate in the \href{https://eval.ai/challenge/1541/overview}{eProduct} competition 
in conjunction with 
the \href{https://sites.google.com/view/fgvc9/home}{FGVC9} at \href{https://cvpr2022.thecvf.com/}{CVPR 2022}.

In this competition, 
we aim to develop an \textit{open-world} retrieval system  generalizable to daily added new products.
We appreciate a couple of key characteristics.
Firstly, product retrieval is essentially a \textit{super fine-grained} task for reflecting individual user's needs, \eg, a buyer needs to find a desired product among a large number of options without easily distinguishable nuances.
Secondly, unlike many popular academic datasets \cite{hadi2015buy,liu2016deepfashion,wu2021fashioniq},
instance-level ground truth labels are much harder to acquire since sellers tend to upload their products independently.
Further, sellers would most likely upload a product with some inaccurate short description and select a category for that item.
As a result, coarse-grained class labels and text descriptions
can be only considered as {\em weak supervision}.

There have been a recent surge in weakly supervised representation learning \cite{zhai2018strongbaseline,boudiaf2020unifying,khosla2020supcon,zhang2022use,tian2020cmc,radford2021clip}.
Several works \cite{zhai2018strongbaseline,boudiaf2020unifying} have empirically and theoretically demonstrated that optimizing the classification objective is the same or even better than optimizing pair-wise objectives for metric learning. 
This implies that a retrieval problem can be simplified into 
a classification problem by predicting coarse-grained class labels.
A classification objective could be typically implemented with cross-entropy loss or supervised contrastive loss \cite{khosla2020supcon,zhang2022use}.
Alternatively, cross-modal contrastive learning \cite{tian2020cmc,radford2021clip} can also 
%
learn discriminative representations from different views (\eg, product image + product titles).
However, we find that all these methods are not competent in the eProduct benchmark due to its aforementioned challenges (also see our evaluation in Section \ref{sec:results}).
We summarize the reasons as follows:
(a) The model cannot learn \textit{instance-level discriminative} features only with a typical classification loss using category-level labels;
(b) The model cannot learn \textit{intra-modal discriminative} features with the cross-modal contrastive loss only.
This motivates us to formulate an {\bf\em intra-modal learning objective} that maximizes the use of weakly-supervised information available.

Inspired by previous works \cite{joulin2016weak1,mahajan2018weak2,singh2022weak3},
we introduce a novel pseudo multi-labeling strategy.
Concretely, we extract pseudo-attributes from product titles and optimize a multi-label classification objective.
This forms a weakly-supervised representation learning problem. 
The advantage of our formulation is two-folds:
(a) 
Our pseudo-attributes \textit{instance-dependent} and thus more fine-grained than category labels, despite being noisy by nature;
(b) 
Our pseudo-attributes retain the majority of information from the text modality, whilst the visual feature learning is conducted in an \textit{intra-modal} manner.
This decoupling design well mitigates the challenges as discussed earlier.
Further, we build a strong baseline model 
with ImageNet-pretrained backbones \cite{liu2021swin,liu2022convnext}, advanced training recipes (\eg, TrivalAugment \cite{muller2021trivialaugment} and random erasing \cite{zhong2020randomerasing}) and post-processing (feature whitening \cite{su2021whitening,huang2021whiteningbert}, k-reciprocal reranking \cite{zhong2017rerank} and model ensemble). 
Empirically, we show that our method outperforms all competing methods by a large margin, getting hold of the {\bf \em second} position out of about twenty teams on the leaderboard.

\section{Methodology}
\label{sec:methodology}
In this section, we describe our solution for this challenge.
In Sec~\ref{subsec:opjectives}, we describe our proposed label-smoothed multi-label classification objective based on pseudo attributes. In Sec~\ref{subsec:recipes}, we discuss various training recipes that we have integrated to make our solution competitive. In Sec~\ref{subsec:postprocessing}, the post-processing techniques are elaborated.  

\subsection{Optimization objectives}
\label{subsec:opjectives}
As described in Section~\ref{sec:intro}, we empirically found that using the available weakly-supervised information is non-trivial for this task.
This is due to the inherent challenge of this task: lack of instance-level fine-grained ground truth.
To circumvent the absence of instance-level labels, we propose to mine pseudo-attributes from the product titles.
Although some different products may share some overlapping pseudo-attributes, the set of attributes for each product is instance-dependent.
Therefore, it is reasonable that the model can learn instance-wise discriminative features from these pseudo-attributes.

\begin{figure*}[ht]
\begin{center}
\includegraphics[width=\linewidth]{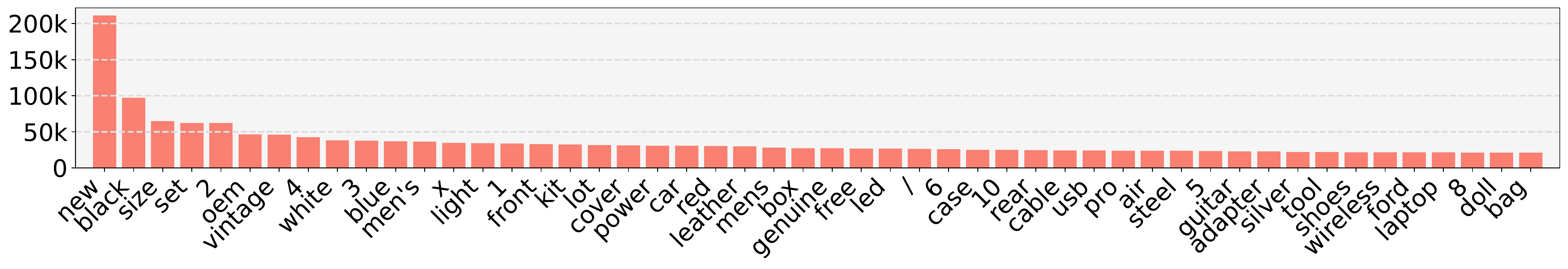}
\includegraphics[width=\linewidth]{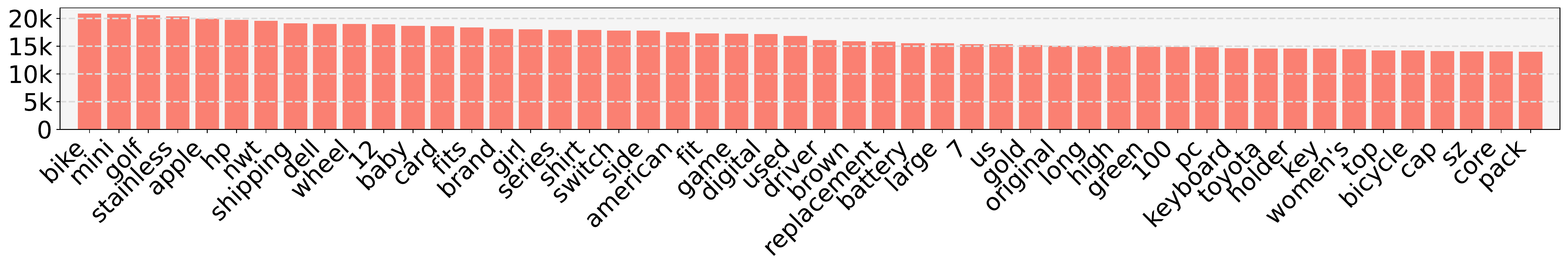}
\end{center}
\vspace{-4mm}
\caption{\textbf{Histogram of the top-100 pseudo attributes mined from product titles.}}
\label{fig:attribute_histogram}
\end{figure*}

Specifically, we extract all words with a simple blank-based tokenizer and only keep the words that appear more than 30 times to filter some noisy words, resulting in a list of 22,295 pseudo-attributes.
We show the histogram of the top-100 pseudo-attributes in Figure~\ref{fig:attribute_histogram}.
It is observed that all of them are truly highly-related to the product information but very noisy. To utilize these mined pseudo-attributes, we then perform multi-label classification for the weakly-supervised representation learning. 
A common trend to solve multi-label classification problem is to use  \texttt{sigmoid} activation followed by logistic regression based loss terms. However, doing so may result in designing  sophisticated label completion techniques and more hyper-parameter search which may reduce the efficacy of the solution. Following the works in \cite{hoe2021orthohash,joulin2016weak1,mahajan2018weak2,singh2022weak3,han2022fashionvil}, we use \texttt{softmax} activation coupled with cross-entropy loss as our optimization objective to overcome the above drawbacks. Formally, it is defined as follows :
%
\begin{align}
    \mathcal{L}_{CE} = - \frac{1}{N} \sum^N_i \sum^T_t Y^{(i)}_{t} \log P^{(i)}_{t},
\end{align}
where $N$ is the number of training data, $T$ is the number of pseudo classes, $Y^{(i)}_t$ is the target which is either $\frac{1}{K}$ or $0$ at class $t$ depending on whether the corresponding pseudo-attribute is present or not. $K$ is the number of pseudo-attributes for $i$-th sample, and $P^{(i)}_t$ is the \texttt{softmax} probability of class $t$.
Due to the long-tail nature of the dataset (ref to Fig. \ref{fig:attribute_histogram}), generating pseudo-attributes may result in severe class-imbalance particularly for the tail attributes. To address this class-imbalance problem, we can apply either FocalLoss \cite{lin2017focalloss} or PolyLoss \cite{leng2022polyloss} as our training objective. For the simplicity in loss design, we chose PolyLoss as the objective function. It is defined as follows :
%
\begin{align} \label{eq:polyloss}
    \mathcal{L}_{poly} = \mathcal{L}_{CE} + \frac{\epsilon}{N} \sum^N_i \sum^T_t (1 - Y^{(i)}_t P^{(i)}_t),
\end{align}
where $\epsilon$ is the scale of the regularization effect of the PolyLoss. We train our model using Eq. \ref{eq:polyloss} with $\epsilon=0.5$. 

\subsection{Training recipes}
\label{subsec:recipes}
\begin{table}[t]
\resizebox{\linewidth}{!}{
\begin{tabular}{c|c}
\textbf{Training config} & \textbf{ConvNeXt-XL\cite{liu2022convnext} / Swin-L\cite{liu2021swin}} \\ 
\shline
training resolution & 224 / 384 \\
inference resoultion & 316 \& 384 / 384 \\
optimizer & AdamW \\
base learning rate & 1e-4\\
weight decay & 1e-4  \\
poly loss $\epsilon$ \cite{leng2022polyloss} & 0.5 \\
optimizer momentum & $\beta_1, \beta_2 = 0.9, 0.999$  \\
batch size & 28 $\times$ 8 / 14 $\times$ 8\\
training epochs & 20  \\
learning rate schedule & cosine decay \\
warmup epochs & 5  \\
warmup schedule & linear  \\
TrivialAugment \cite{muller2021trivialaugment} & \textit{num\_magnitude\_bins} = 31  \\
random erasing \cite{zhong2020randomerasing} & 0.25  \\
stochastic depth \cite{huang2016stochasticdepth} & 0.4  \\
head init scale \cite{touvron2021goingdeeper} & 0.01  \\
exp. mov. avg. (EMA) \cite{polyak1992ema} & 0.9999 \\
reranking \cite{zhong2017rerank} & $K_1,K_2,\alpha=8,5,0.5$\\
\end{tabular}
}
\caption{\textbf{All hyper-parameters used in our solution}. Multiple hyper-parameters (\eg, 224 / 384 training resolution) are for each model (\eg, ConvNeXt-XL / Swin-L) respectively.}
\label{tab:train_detail}
\end{table}
We used the recently proposed CNN-based ConvNeXt \cite{liu2022convnext} and Transformer-based Swin-L \cite{liu2021swin} as our model backbones followed by a linear projection and a \texttt{softmax} activation is added after the output from the backbone
to compute the prediction probability $P$ in Eq. \ref{eq:polyloss} respectively. 

For all other training configurations, including tricks (\eg, TrivialAugment \cite{muller2021trivialaugment}, and random erasing \cite{zhong2020randomerasing}) and hyper-parameters (\eg, batch size, learning rate, and learning scheduler), we have summarized them in Table \ref{tab:train_detail}.

\subsection{Post-processing}
\label{subsec:postprocessing}
\noindent \textbf{Inference resolution.} 
To extract features for retrieval, we compute using the image resolution of $316^2$/$384^2$ for ConvNeXt and $384^2$ for Swin-L. It is because larger image resolution can extract finer information from an image. 

\statement{Feature whitening.} To reduce correlation between dimensions of the features that could possibly improve the retrieval results \cite{su2021whitening,huang2021whiteningbert}, we further apply feature whitening on both query and database features. 
The parameters used by feature whitening are computed using database features only. Finally, the whitened features are $l_2$ normalized. 

\statement{Ensemble models.} Ensembling can make better predictions and achieve better performance as it reduces the spread of the predictions (\eg, a single model might overfit or bias to some certain patterns, ensemble model can reduce the effect). 
We ensemble a ConvNeXt-XL model and three Swin-L models trained with different random seeds and slightly different learning rates.
Note that feature whitening and $l_2$ normalization are applied for each model independently. All features are then concatenated at feature dimension.

\statement{Distance metric.} During retrieval, we use cosine distance as the distance metric. We have also experimented with Euclidean distance, but we empirically found out that cosine distance has consistently 1-2\% MAR@10 improvement.

\statement{\textit{k}-reciprocal re-ranking.} 
After retrieving top-$N$ items from the index database, we then performed \textit{k}-reciprocal re-ranking \cite{zhong2017rerank} among the top-$N$ items. 
We observed that re-ranking always boosted the MAR@10 for at least 4\% in absolute performance. 

\statement{Algorithm.} We have summarized our retrieval pipeline with Pytorch-style pseudocode in Algorithm \ref{alg:pipeline}.

\begin{algorithm}
\caption{Pseudocode of our retrieval pipeline.}
\label{alg:pipeline}
\definecolor{codeblue}{rgb}{0.25,0.5,0.5}
\lstset{
  backgroundcolor=\color{white},
  basicstyle=\fontsize{7.2pt}{7.2pt}\ttfamily\selectfont\color{codeblue},
  columns=fullflexible,
  breaklines=true,
  captionpos=b,
  commentstyle=\fontsize{7.2pt}{7.2pt},
  keywordstyle=\fontsize{7.2pt}{7.2pt}\color{blue},
}
\begin{lstlisting}[language=python]
def compute_feats(model):
    q_feats = model(q_imgs)  # (N, d)
    db_feats = model(db_imgs)  # (M, d)
    
    db_mean, db_cov = compute_mean_cov(db_feats)
    
    q_feats = whitening(q_feats, db_mean, db_cov)
    db_feats = whitening(db_feats, db_mean, db_cov)
    
    return l2_norm(q_feats), l2_norm(db_feats)
    
q_feats_ens = []
db_feats_ens = []
for model in ensemble:
    q_feats, db_feats = compute_feats(model)
    q_feats_ens.append(q_feats)
    db_feats_ens.append(db_feats)

q_feats_ens = concat(q_feats_ens, dim=1)
db_feats_ens = concat(db_feats_ens, dim=1)

topn_db_feats = retrieve(q_feats_ens, db_feats_ens)
top10_db_ids = reranking(q_feats_ens, topn_db_feats)
\end{lstlisting}
\end{algorithm}

\section{Experiments}
\subsection{Implementation details}
All of our experiments are performed using 8 NVIDIA GeForce RTX 3090 GPUs. 
We use the validation set for model evaluation and we exclude the validation split images from training.
All the hyper-parameters used in our training recipes are listed in Table~\ref{tab:train_detail}.
We use Pytorch Lightning \cite{falcon2019pl} as our framework and use Hydra \cite{yadan2019hydra} as configuration system.
We have used the ImageNet pre-trained ConvNext-XL and Swin-L models provided by \texttt{timm} \cite{rw2019timm}. 

\begin{table}[t]
\centering
\resizebox{0.9\linewidth}{!}{
\begin{tabular}{c|c|c|c}
    \textbf{Method} & \textbf{Used Info.} & \textbf{Objective} & \textbf{MAR@10} \\
    \shline
    SLC & coarse labels & SupCon~\cite{khosla2020supcon}  & 16.47 \\
    SLC & coarse labels & X-Entropy~\cite{zhai2018strongbaseline,boudiaf2020unifying} & 21.61 \\
    CMC & product titles & InfoNCE~\cite{tian2020cmc,radford2021clip} & 32.59 \\
    SLC & pseudo labels & SupCon~\cite{caron2018deepclustering} & 37.35 \\
    MLC & pseudo attributes & X-Entropy & \textbf{45.69} \\
\end{tabular}
}
\caption{\textbf{The comparisons of different learning objectives.}
SLC: Single-Label Classification / Contrastive;
CMC: Contratsive Multi-view Coding;
MLC: Multi-Label Classification.}
\label{tab:method_ablation}
\end{table}
\subsection{Evaluation metric}
For model evaluation, macro-average recall@k (\ie, MAR@k) metric is used.
It is defined as the average of recall@k over all $N$ queries: $\textbf{MAR}@k=\frac{1}{N} \sum^N_{i=1} \textbf{R}_i$,
where $k$ is the number of top-$k$ retrieved items, and $\textbf{R}_i$ is the recall@k of $i$-th query.
Quantitatively, if this metric is higher, it is better.
Following the evaluation protocol of eProduct~\cite{yuan2021eproduct}, we use MAR@10 as the evaluation metric, where $k=10$ and $N=3000$ for test phase.

\subsection{Results}
\label{sec:results}
In this subsection, we show the effectiveness of our proposed learning objective and different modules mentioned in Section~\ref{sec:methodology}. 
Note that we can only report \textit{approximate} numbers here, as we no longer have access to ground truth labels for accurate evaluation.

\statement{Comparison with competing methods.}
To demonstrate the superiority of our proposed multi-label classification optimization objective, we conducted several comparative experiments with other alternative methods based on ResNet50~\cite{he2016resnet}.
We consider several methods including:
\begin{itemize}[itemsep=0pt,topsep=1pt,parsep=0pt]
    \item Use the coarse-grained category labels as the targets and then optimize the model as single-label classification. Here, we consider two objectives: cross-entropy (X-Entropy)~\cite{zhai2018strongbaseline,boudiaf2020unifying} and supervised contrastive (SupCon)~\cite{khosla2020supcon}.
    This method has been proved to be effective for coarse-grained image retrieval~\cite{zhai2018strongbaseline,boudiaf2020unifying}.
    \item Use the product titles as the natural augmentation view of the product images and then optimize the model via cross-modal contratsive learning~\cite{tian2020cmc}. 
    This method achieves great success in weakly-supervised cross-modal representation learning~\cite{radford2021clip}.
    \item Use off-the-shelf traditional clustering method (\eg, DBSCAN~\cite{ester1996dbscan}) to assign pseudo instance labels for every image based on its feature and then use these pseudo labels as the ground truth.
    This method can be regarded as a special case of unsupervised cluster-based representation learning~\cite{caron2018deepclustering,chen2021ice}.
\end{itemize}

We show the performance comparison in Table~\ref{tab:method_ablation}.
It is obvious that our multi-label classification outperforms all other methods by a large margin.
We conclude that this superiority can be attributed to two advantages of our learning objective:
(a) the mined pseudo-attributes are instance-dependent and thus more fine-grained than category labels;
(b) our learning process is intra-modal, thereby making the learned image features directly comparable to others.
More interestingly, we observe that the clustering based single-label unsupervised contrastive learning approach outperforms all the other weakly-supervised methods except ours.
This phenomenon reflects on the fact that  it is non-trivial to use weakly-supervised information
Hence, if the learning objective is not designed properly, the model tends to overfit the coarse-grained information.

\statement{Ablation study of engineering tricks.}
We show the ablation study of different components of our proposed solution in Table~\ref{tab:component_ablation}.
We empirically found that stochastic depth~\cite{huang2016stochasticdepth} can boost the performance a lot by alleviating the process of over-fitting.
We also found that whitening~\cite{su2021whitening,huang2021whiteningbert} and \textit{k}-reciprocal re-ranking~\cite{zhong2017rerank} are highly necessary for this open-world retrieval task.

\begin{table}[t]
\centering
\resizebox{0.7\linewidth}{!}{
\begin{tabular}{l|c}
    \multicolumn{1}{c|}{\textbf{Component}} & \textbf{MAR@10} \\
    \shline
    Baseline (ConvNeXt-XL~\cite{liu2022convnext}) & $\sim$ 59.00 \\
    + Higher Inference Resolution & $\sim$ 61.00 \\
    + Drop Path & $\sim$ 64.00 \\
    + PolyLoss ($\epsilon=0.5$) & $\sim$ 64.50 \\
    + EMA & $\sim$ 65.00 \\
    + Feature Whitening & $\sim$ 66.00 \\
    + Re-ranking & $\sim$ 69.00 \\
    + Ensemble (4 models) & \textbf{71.53} \\
\end{tabular}
}
\caption{\textbf{The ablation study of different components we used.} The symbol $\sim$ represents approximate numbers.}
\label{tab:component_ablation}
\end{table}
\section{Conclusion}
In this challenge, we introduce a pseudo multi-labeling strategy.
We mine pseudo-attributes from product titles for weakly supervised representation learning. 
Given many products with subtle visual differences, the mined pseudo-attributes act as an instance-level label for enabling instance-level discriminative feature learning. By applying strong backbones, advanced training recipes and post-processing techniques, our solution achieves 71.53\% MAR@10 and ranks the {\bf second} on the leaderboard of eBay Visual Search Challenge in FGVC9, CVPR 2022. Our take-home message is that, weakly supervised representation learning with instance-level pseudo-attributes is particularly powerful. We hope that this finding can foster more related research in academia and industry. 

\section*{Acknowledgement}
We would like to thank Dr. Xiatian Zhu for help with editing the manuscript.
Additionally, we would like to thank Dr. Li Zhang, Prof. Yi-Zhe Song and Prof. Tao Xiang for their support of this competition. 

{\small
\bibliographystyle{ieee_fullname}
\bibliography{egbib}

\begin{thebibliography}{10}\itemsep=-1pt

\bibitem{boudiaf2020unifying}
Malik Boudiaf, J{\'e}r{\^o}me Rony, Imtiaz~Masud Ziko, Eric Granger, Marco
  Pedersoli, Pablo Piantanida, and Ismail~Ben Ayed.
\newblock A unifying mutual information view of metric learning: cross-entropy
  vs. pairwise losses.
\newblock In {\em ECCV}, 2020.

\bibitem{caron2018deepclustering}
Mathilde Caron, Piotr Bojanowski, Armand Joulin, and Matthijs Douze.
\newblock Deep clustering for unsupervised learning of visual features.
\newblock In {\em ECCV}, 2018.

\bibitem{chen2021ice}
Hao Chen, Benoit Lagadec, and Francois Bremond.
\newblock Ice: Inter-instance contrastive encoding for unsupervised person
  re-identification.
\newblock In {\em ICCV}, 2021.

\bibitem{ester1996dbscan}
Martin Ester, Hans-Peter Kriegel, J{\"o}rg Sander, Xiaowei Xu, et~al.
\newblock A density-based algorithm for discovering clusters in large spatial
  databases with noise.
\newblock In {\em KDD}, 1996.

\bibitem{falcon2019pl}
William Falcon et~al.
\newblock Pytorch lightning.
\newblock Github, 2019.
\newblock \url{https://github.com/PyTorchLightning/pytorch-lightning}.

\bibitem{hadi2015buy}
M Hadi~Kiapour, Xufeng Han, Svetlana Lazebnik, Alexander~C Berg, and Tamara~L
  Berg.
\newblock Where to buy it: Matching street clothing photos in online shops.
\newblock In {\em ICCV}, 2015.

\bibitem{han2022fashionvil}
Xiao Han, Licheng Yu, Xiatian Zhu, Li Zhang, Yi-Zhe Song, and Tao Xiang.
\newblock Fashionvil: Fashion-focused vision-and-language representation
  learning.
\newblock In {\em ECCV}, 2022.

\bibitem{he2016resnet}
Kaiming He, Xiangyu Zhang, Shaoqing Ren, and Jian Sun.
\newblock Deep residual learning for image recognition.
\newblock In {\em CVPR}, 2016.

\bibitem{hoe2021orthohash}
Jiun~Tian Hoe, Kam~Woh Ng, Tianyu Zhang, Chee~Seng Chan, Yi-Zhe Song, and Tao
  Xiang.
\newblock One loss for all: Deep hashing with a single cosine similarity based
  learning objective.
\newblock {\em NeurIPS}, 2021.

\bibitem{huang2016stochasticdepth}
Gao Huang, Yu Sun, Zhuang Liu, Daniel Sedra, and Kilian~Q Weinberger.
\newblock Deep networks with stochastic depth.
\newblock In {\em ECCV}, 2016.

\bibitem{huang2021whiteningbert}
Junjie Huang, Duyu Tang, Wanjun Zhong, Shuai Lu, Linjun Shou, Ming Gong, Daxin
  Jiang, and Nan Duan.
\newblock Whiteningbert: An easy unsupervised sentence embedding approach.
\newblock In {\em EMNLP}, 2021.

\bibitem{joulin2016weak1}
Armand Joulin, Laurens van~der Maaten, Allan Jabri, and Nicolas Vasilache.
\newblock Learning visual features from large weakly supervised data.
\newblock In {\em ECCV}, 2016.

\bibitem{khosla2020supcon}
Prannay Khosla, Piotr Teterwak, Chen Wang, Aaron Sarna, Yonglong Tian, Phillip
  Isola, Aaron Maschinot, Ce Liu, and Dilip Krishnan.
\newblock Supervised contrastive learning.
\newblock {\em NeurIPS}, 2020.

\bibitem{leng2022polyloss}
Zhaoqi Leng, Mingxing Tan, Chenxi Liu, Ekin~Dogus Cubuk, Jay Shi, Shuyang
  Cheng, and Dragomir Anguelov.
\newblock Polyloss: A polynomial expansion perspective of classification loss
  functions.
\newblock In {\em ICLR}, 2022.

\bibitem{lin2017focalloss}
Tsung-Yi Lin, Priya Goyal, Ross Girshick, Kaiming He, and Piotr Doll{\'a}r.
\newblock Focal loss for dense object detection.
\newblock In {\em ICCV}, 2017.

\bibitem{liu2021swin}
Ze Liu, Yutong Lin, Yue Cao, Han Hu, Yixuan Wei, Zheng Zhang, Stephen Lin, and
  Baining Guo.
\newblock Swin transformer: Hierarchical vision transformer using shifted
  windows.
\newblock In {\em ICCV}, 2021.

\bibitem{liu2016deepfashion}
Ziwei Liu, Ping Luo, Shi Qiu, Xiaogang Wang, and Xiaoou Tang.
\newblock Deepfashion: Powering robust clothes recognition and retrieval with
  rich annotations.
\newblock In {\em CVPR}, 2016.

\bibitem{liu2022convnext}
Zhuang Liu, Hanzi Mao, Chao-Yuan Wu, Christoph Feichtenhofer, Trevor Darrell,
  and Saining Xie.
\newblock A convnet for the 2020s.
\newblock {\em arXiv preprint arXiv:2201.03545}, 2022.

\bibitem{mahajan2018weak2}
Dhruv Mahajan, Ross Girshick, Vignesh Ramanathan, Kaiming He, Manohar Paluri,
  Yixuan Li, Ashwin Bharambe, and Laurens Van Der~Maaten.
\newblock Exploring the limits of weakly supervised pretraining.
\newblock In {\em ECCV}, 2018.

\bibitem{muller2021trivialaugment}
Samuel~G M{\"u}ller and Frank Hutter.
\newblock Trivialaugment: Tuning-free yet state-of-the-art data augmentation.
\newblock In {\em ICCV}, 2021.

\bibitem{polyak1992ema}
Boris~T Polyak and Anatoli~B Juditsky.
\newblock Acceleration of stochastic approximation by averaging.
\newblock {\em SIAM journal on control and optimization}, 1992.

\bibitem{radford2021clip}
Alec Radford, Jong~Wook Kim, Chris Hallacy, Aditya Ramesh, Gabriel Goh,
  Sandhini Agarwal, Girish Sastry, Amanda Askell, Pamela Mishkin, Jack Clark,
  et~al.
\newblock Learning transferable visual models from natural language
  supervision.
\newblock In {\em ICML}, 2021.

\bibitem{singh2022weak3}
Mannat Singh, Laura Gustafson, Aaron Adcock, Vinicius de~Freitas Reis, Bugra
  Gedik, Raj~Prateek Kosaraju, Dhruv Mahajan, Ross Girshick, Piotr Doll{\'a}r,
  and Laurens van~der Maaten.
\newblock Revisiting weakly supervised pre-training of visual perception
  models.
\newblock In {\em CVPR}, 2022.

\bibitem{su2021whitening}
Jianlin Su, Jiarun Cao, Weijie Liu, and Yangyiwen Ou.
\newblock Whitening sentence representations for better semantics and faster
  retrieval.
\newblock {\em arXiv preprint arXiv:2103.15316}, 2021.

\bibitem{tian2020cmc}
Yonglong Tian, Dilip Krishnan, and Phillip Isola.
\newblock Contrastive multiview coding.
\newblock In {\em ECCV}, 2020.

\bibitem{touvron2021goingdeeper}
Hugo Touvron, Matthieu Cord, Alexandre Sablayrolles, Gabriel Synnaeve, and
  Herv{\'e} J{\'e}gou.
\newblock Going deeper with image transformers.
\newblock In {\em ICCV}, 2021.

\bibitem{rw2019timm}
Ross Wightman.
\newblock Pytorch image models.
\newblock GitHub, 2019.
\newblock \url{https://github.com/rwightman/pytorch-image-models}.

\bibitem{wu2021fashioniq}
Hui Wu, Yupeng Gao, Xiaoxiao Guo, Ziad Al-Halah, Steven Rennie, Kristen
  Grauman, and Rogerio Feris.
\newblock Fashion iq: A new dataset towards retrieving images by natural
  language feedback.
\newblock In {\em CVPR}, 2021.

\bibitem{yadan2019hydra}
Omry Yadan.
\newblock Hydra - a framework for elegantly configuring complex applications.
\newblock Github, 2019.
\newblock \url{https://github.com/facebookresearch/hydra}.

\bibitem{yu2022commercemm}
Licheng Yu, Jun Chen, Animesh Sinha, Mengjiao~MJ Wang, Hugo Chen, Tamara~L
  Berg, and Ning Zhang.
\newblock Commercemm: Large-scale commerce multimodal representation learning
  with omni retrieval.
\newblock In {\em KDD}, 2022.

\bibitem{yuan2021eproduct}
Jiangbo Yuan, An-Ti Chiang, Wen Tang, and Antonio Haro.
\newblock eproduct: A million-scale visual search benchmark to address product
  recognition challenges.
\newblock {\em arXiv preprint arXiv:2107.05856}, 2021.

\bibitem{zhai2018strongbaseline}
Andrew Zhai and Hao-Yu Wu.
\newblock Classification is a strong baseline for deep metric learning.
\newblock In {\em BMVC}, 2018.

\bibitem{zhang2022use}
Shu Zhang, Ran Xu, Caiming Xiong, and Chetan Ramaiah.
\newblock Use all the labels: A hierarchical multi-label contrastive learning
  framework.
\newblock In {\em CVPR}, 2022.

\bibitem{zhong2017rerank}
Zhun Zhong, Liang Zheng, Donglin Cao, and Shaozi Li.
\newblock Re-ranking person re-identification with k-reciprocal encoding.
\newblock In {\em CVPR}, 2017.

\bibitem{zhong2020randomerasing}
Zhun Zhong, Liang Zheng, Guoliang Kang, Shaozi Li, and Yi Yang.
\newblock Random erasing data augmentation.
\newblock In {\em AAAI}, 2020.

\end{thebibliography}
}

\end{document}